\documentstyle[colacl,epsfig]{article}

\title{Exploiting Syntactic Structure for Language Modeling}
\author{Ciprian Chelba \and Frederick Jelinek\\Center for Language and
  Speech Processing\\ The Johns Hopkins University, Barton Hall 320\\
  3400 N. Charles St., Baltimore, MD-21218, USA\\
  \{chelba,jelinek\}@jhu.edu
}

\begin{document}
\maketitle

\begin{abstract}%
The paper presents a language model that develops syntactic structure
and uses it to extract meaningful information from the word
history, thus enabling the use of long distance dependencies. The
model assigns probability to every joint sequence of
words--binary-parse-structure with headword annotation and operates in
a left-to-right manner --- therefore usable for automatic speech
recognition. The model, its probabilistic parameterization, and a set of experiments
meant to evaluate its predictive power are presented; an improvement
over standard trigram modeling is achieved.
\end{abstract}

\section{Introduction}

The main goal of the present work is to develop a language model that
uses syntactic structure to model long-distance dependencies. During
the summer96 DoD Workshop a similar attempt was made by the dependency
modeling group. The model we present is closely related to the one
investigated in~\cite{ws96}, however different in a few important
aspects:\\
$\bullet$ our model operates in a left-to-right manner, allowing the
  decoding of word lattices, as opposed to the one referred to
  previously, where only whole sentences could be processed, thus
  reducing its applicability to n-best list re-scoring; the syntactic
  structure is developed as a model component; \\
$\bullet$ our model is a factored version of the one in~\cite{ws96}, thus enabling the
  calculation of the joint probability of words and parse structure;
  this was not possible in the previous case due to the huge
  computational complexity of the model.

Our model develops syntactic structure incrementally while traversing
the sentence from left to right. This is the main
difference between our approach and other approaches to statistical
natural language parsing. Our parsing strategy is similar to the incremental
syntax ones proposed relatively recently in the linguistic community~\cite{colin96}. 
The probabilistic model, its parameterization and a few experiments
that are meant to evaluate its potential for speech recognition are
presented. 

\section{The Basic Idea and Terminology} \label{section:basic_idea}

\begin{figure}
  \begin{center} 
    \epsfig{file=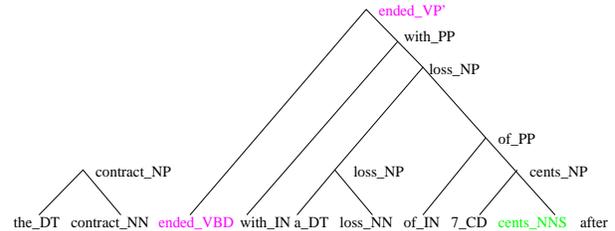,height=3cm,width=\columnwidth}
  \end{center}
  \caption{Partial parse} \label{fig:p_parse}
\end{figure}

Consider predicting the word \verb+after+ in the sentence:\\
\verb+the contract ended with a loss of 7 cents+ \\ \verb+after trading as low as 89 cents+.\\
A 3-gram approach would predict \verb+after+ from \verb+(7, cents)+
whereas it is intuitively clear that the strongest predictor would be
\verb+ended+ which is outside the reach of even 7-grams.
Our assumption is that what enables humans to make a
good prediction of \verb+after+ is the syntactic structure in the
past. The linguistically correct \emph{partial parse} of
the word history when predicting \verb+after+ is shown in
Figure~\ref{fig:p_parse}. 
The word \verb+ended+ is called the \emph{headword} of the
\emph{constituent} \verb+(ended (with (...)))+
and \verb+ended+ is an \emph{exposed headword} when predicting
\verb+after+ --- topmost headword in the largest constituent that
contains it. The syntactic structure in the past filters out
irrelevant words and points to the important ones, thus enabling the
use of long distance information when predicting the next word. 

Our model will attempt to build the syntactic structure incrementally
while traversing the sentence left-to-right. 
The model will assign a probability $P(W,T)$ to every sentence $W$
with every possible POStag assignment, binary branching parse, non-terminal label
and headword annotation for every constituent of $T$.

\begin{figure} 
  \begin{center}
    \epsfig{file=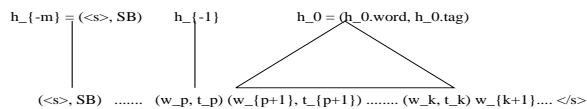,height=1.3cm,width=\columnwidth}
  \end{center}
  \caption{A word-parse k-prefix} \label{fig:w_parse}
\end{figure}

\begin{figure} 
  \begin{center} 
    \epsfig{file=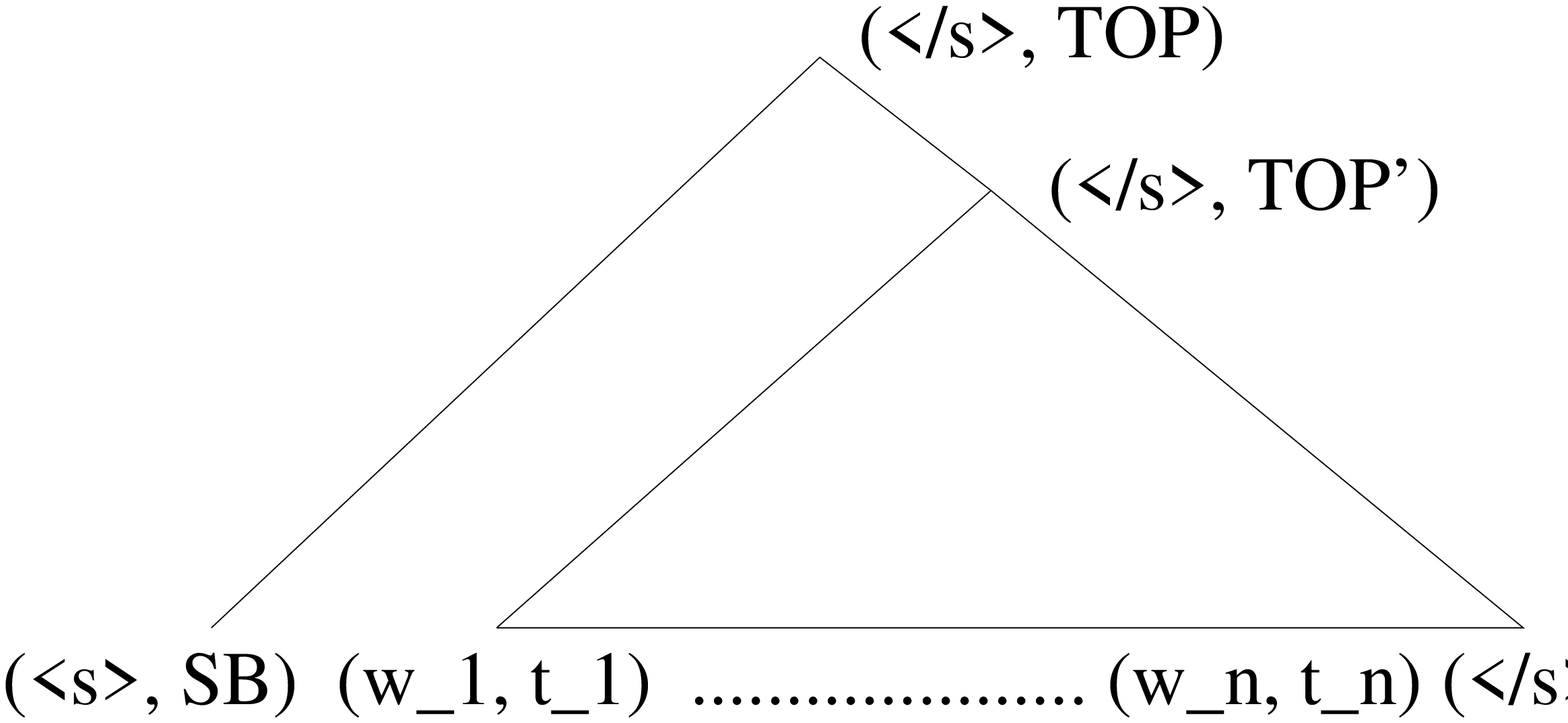,height=2cm,width=4cm}
  \end{center}
  \caption{Complete parse} \label{fig:c_parse}
\end{figure}

Let $W$ be a sentence of length $n$ words to which we have prepended
\verb+<s>+ and appended \verb+</s>+ so that $w_0 = $\verb+<s>+ and
$w_{n+1} = $\verb+</s>+.
Let $W_k$ be the word k-prefix $w_0 \ldots w_k$ of the sentence and 
\mbox{$W_k T_k$} the \emph{word-parse k-prefix}. To stress this point, a
\mbox{word-parse k-prefix} contains --- for a given parse --- only those binary
subtrees whose span is completely included in the word k-prefix, excluding 
$w_0 = $\verb+<s>+. Single words along with their POStag can be
regarded as root-only trees. Figure~\ref{fig:w_parse} shows a
word-parse k-prefix; \verb|h_0 .. h_{-m}| are the \emph{exposed
 heads}, each head being a pair(headword, non-terminal label), or
(word,  POStag) in the case of a root-only tree. 
A \emph{complete parse} --- Figure~\ref{fig:c_parse} --- is any binary
parse of the \\ \mbox{$(w_1,t_1)  \ldots (w_n,t_n)$\verb+ (</s>, SE)+}
sequence with the restriction that \verb+(</s>, TOP')+ is the only
allowed head. Note that 
\mbox{$((w_1,t_1) \ldots (w_n,t_n))$} \emph{needn't} be a constituent,
but for the parses where it is, there is no restriction on which of
its words is the headword or what is the non-terminal label that
accompanies the headword. 

The model will operate by means of three modules:\\
$\bullet$ WORD-PREDICTOR predicts the next word $w_{k+1}$ given the
  word-parse k-prefix and then passes control to the TAGGER;\\
$\bullet$ TAGGER predicts the POStag of the next word $t_{k+1}$ given the
  word-parse k-prefix and the newly predicted word and then passes
  control to the PARSER;\\
$\bullet$ PARSER grows the already existing binary branching structure by
  repeatedly generating the transitions:\\ \verb+(unary, NTlabel)+,
  \verb+(adjoin-left, NTlabel)+ or  \verb+(adjoin-right, NTlabel)+
  until it passes control to the PREDICTOR
  by  taking a \verb+null+ transition. \verb+NTlabel+ is the non-terminal
  label assigned to the newly built constituent and
  \verb+{left,right}+ specifies where the new headword is inherited
  from.

\begin{figure}
  \begin{center} 
    \epsfig{file=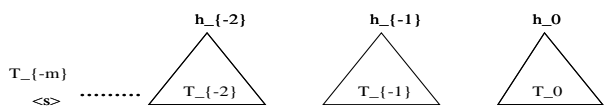,height=1.3cm,width=\columnwidth}
  \end{center}
  \caption{Before an adjoin operation} \label{fig:before}
\end{figure}

\begin{figure}
  \begin{center} 
    \epsfig{file=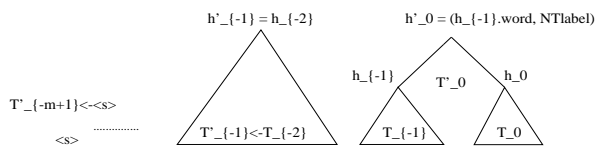,height=1.8cm,width=\columnwidth}
  \end{center}
  \caption{Result of adjoin-left under NTlabel} \label{fig:after_a_l}
\end{figure}

\begin{figure}
  \begin{center} 
    \epsfig{file=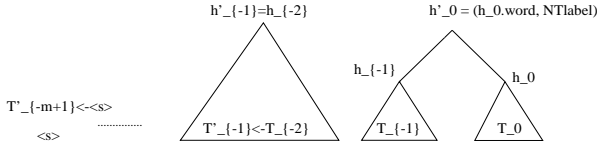,height=2cm,width=\columnwidth}
  \end{center}
  \caption{Result of adjoin-right under NTlabel} \label{fig:after_a_r}
\end{figure}

The operations performed by the PARSER are illustrated in
Figures~\ref{fig:before}-\ref{fig:after_a_r} and they ensure that all possible binary
branching parses with all possible headword and non-terminal label
assignments for the $w_1 \ldots w_k$ word sequence can be generated.
The following algorithm formalizes the above description of the
sequential generation of a sentence with a complete parse.
\begin{verbatim}
Transition t;         // a PARSER transition
predict (<s>, SB);
do{
  //WORD-PREDICTOR and TAGGER
  predict (next_word, POStag);   
  //PARSER
  do{                            
    if(h_{-1}.word != <s>){
      if(h_0.word == </s>)
        t = (adjoin-right, TOP');
      else{
        if(h_0.tag == NTlabel)
          t = [(adjoin-{left,right}, NTlabel), 
               null];
        else
          t = [(unary, NTlabel), 
               (adjoin-{left,right}, NTlabel), 
               null];
      }
    }
    else{
      if(h_0.tag == NTlabel)
        t = null;
      else
        t = [(unary, NTlabel), null];
    }
  }while(t != null) //done PARSER
}while(!(h_0.word==</s> && h_{-1}.word==<s>))
t = (adjoin-right, TOP); //adjoin <s>_SB; DONE;
\end{verbatim}

The unary transition is allowed only when the most recent exposed
head is a leaf of the tree --- a regular word along with its POStag
--- hence it can be taken at most once at a given position in the
input word string. The second subtree in Figure~\ref{fig:w_parse} provides an example
of a unary transition followed by a null transition.

It is easy to see that any given word sequence with a possible parse
and headword annotation is generated by a unique sequence of model
actions. This will prove very useful in initializing our model
parameters from a treebank --- see
section~\ref{section:initial_parameters}.

\section{Probabilistic Model} \label{section:prob_model}

The probability $P(W,T)$ of a word sequence $W$ and a complete parse
$T$ can be broken into:
\begin{eqnarray}
\lefteqn{P(W,T)= } \nonumber\\
& \prod_{k=1}^{n+1}[&P(w_k/W_{k-1}T_{k-1}) \cdot P(t_k/W_{k-1}T_{k-1},w_k) \cdot \nonumber\\
&  & \prod_{i=1}^{N_k}P(p_i^k/W_{k-1}T_{k-1},w_k,t_k,p_1^k\ldots
p_{i-1}^k)] \label{eq:model}
\end{eqnarray}
where: \\
$\bullet$ $W_{k-1} T_{k-1}$ is the word-parse $(k-1)$-prefix\\
$\bullet$ $w_k$ is the word predicted by WORD-PREDICTOR\\
$\bullet$ $t_k$ is the tag assigned to $w_k$ by the TAGGER\\
$\bullet$ $N_k - 1$ is the number of operations the PARSER executes
  before passing control to the  WORD-PREDICTOR (the $N_k$-th operation at
  position k is the \verb+null+ transition); $N_k$ is a function of $T$\\
$\bullet$ $p_i^k$ denotes the i-th PARSER operation carried out at
  position k in the word string; \\ 
  \mbox{$p_{1}^k \in\{$\verb+(unary, NTlabel)+$,$}
  \mbox{\verb+(adjoin-left, NTlabel)+$,$}\\
  \mbox{\verb+(adjoin-right, NTlabel), null+$\}$,} \\
  \mbox{$p_{i}^k \in\{$}
  \mbox{\verb+(adjoin-left, NTlabel)+$,$}\\
  \mbox{\verb+(adjoin-right, NTlabel)+$\}, 1 < i < N_k$ ,} \\
  \mbox{$p_{i}^k = $\verb+null+$, i = N_k$ } \\

Our model is based on three probabilities:
\begin{eqnarray}
  P(w_k/W_{k-1} T_{k-1})\label{eq:1}\\
  P(t_k/w_k,W_{k-1} T_{k-1})\label{eq:2}\\
  P(p_i^k/w_k,t_k,W_{k-1} T_{k-1}, p_1^k \ldots p_{i-1}^k)\label{eq:3}
\end{eqnarray}
As can be seen, \mbox{$(w_k,t_k,W_{k-1} T_{k-1}, p_1^k \ldots
  p_{i-1}^k)$} is one of the $N_k$ word-parse k-prefixes $W_k T_k$ at
position $k$ in the sentence, $i=\overline{1,N_k}$. 

To ensure a proper probabilistic model (\ref{eq:model}) we have to make sure
that (\ref{eq:1}), (\ref{eq:2}) and (\ref{eq:3}) are well defined conditional
probabilities and that the model halts with probability
one. Consequently, certain PARSER and WORD-PREDICTOR probabilities
must be given specific values:\\
$\bullet$  $P($\verb+null+$/W_kT_k) = 1$, if
\verb+h_{-1}.word = <s>+ and \verb+h_{0}+ $\neq$ \verb+(</s>, TOP')+
--- that is, before predicting \verb+</s>+ --- ensures that \verb+(<s>, SB)+
is adjoined in the last step of the parsing process;\\ 
$\bullet$  \mbox{$P($\verb+(adjoin-right, TOP)+$/W_k T_k) =1$},
if\\ \verb+h_0 = (</s>, TOP')+ and \verb+h_{-1}.word = <s>+ \\and\\
\mbox{$P($\verb+(adjoin-right, TOP')+$/W_k T_k) =1$},
if\\ \verb+h_0 = (</s>, TOP')+ and \verb+h_{-1}.word+ $\neq$ \verb+<s>+ \\
ensure that the parse generated by our model is consistent with the
definition of a complete parse;\\
$\bullet$  $P($\verb+(unary, NTlabel)+$/W_k T_k) =0$,
if \verb+h_0.tag+ $\neq$ \verb+POStag+ ensures correct treatment of unary productions;\\
$\bullet$ \mbox{$\exists\epsilon > 0, \forall W_{k-1}T_{k-1},
  P(w_k$=\verb+</s>+$/W_{k-1}T_{k-1})\geq\epsilon$}\\ ensures that the
model halts with probability one.

The word-predictor model (\ref{eq:1}) predicts the next word based on
the preceding 2 \emph{exposed heads}, thus making the following
equivalence classification: $$P(w_k/W_{k-1} T_{k-1}) = P(w_k/h_0, h_{-1})$$ 

After experimenting with several equivalence classifications of the
word-parse prefix for the tagger model, the conditioning part of model
(\ref{eq:2}) was reduced to using the word to be tagged and the tags
of the two most recent exposed heads: $$P(t_k/w_k,W_{k-1} T_{k-1}) =  P(t_k/w_k, h_0.tag, h_{-1}.tag)$$

Model (\ref{eq:3}) assigns probability to different parses of the word
k-prefix by chaining the elementary operations described above. 
The workings of the parser module are similar to those of Spatter~\cite{spatter}.
The equivalence classification of the \mbox{$W_k T_k$} word-parse we
used for the parser model (\ref{eq:3}) was the same as the one used
in~\cite{mike96}: $$P(p_i^k/W_{k}T_{k}) = P(p_i^k/h_0, h_{-1})$$

It is worth noting that if the binary branching structure
developed by the parser were always right-branching and we mapped the
POStag and non-terminal label vocabularies to a single type then our
model would be equivalent to a trigram language model.

\subsection{Modeling Tools} \label{subsection:modeling_tools}

All model components --- WORD-PREDICTOR, TAGGER, PARSER --- are
conditional probabilistic models of the type $P(y/x_1, x_2, \ldots ,x_n)$ where $y,
x_1, x_2, \ldots ,x_n$ belong to a mixed bag of words, POStags,
non-terminal labels and parser operations ($y$ only). 
For simplicity, the modeling method we chose was deleted
interpolation among relative frequency estimates of different orders $f_n(\cdot)$
using a recursive mixing scheme:
\begin{eqnarray}
\lefteqn{P(y/x_1,\ldots,x_n) =} \nonumber \\ 
&\lambda(x_1,\ldots,x_n) \cdot P(y/x_1,\ldots,x_{n-1}) + \nonumber \\
& (1 - \lambda(x_1, \ldots ,x_n)) \cdot f_n(y/x_1, \ldots ,x_n),\\ 
\lefteqn{f_{-1}(y) = uniform(vocabulary(y))}
\end{eqnarray}
As can be seen, the context mixing scheme discards items in the
context in right-to-left order.
The $\lambda$ coefficients are tied based on the range of the count
$C(x_1, \ldots , x_n)$.  The approach is a standard one which doesn't
require an extensive description given the literature available on
it~\cite{jelinek80}.

\subsection{Search Strategy}

Since the number of parses  for a given word prefix $W_{k}$ grows
exponentially with $k$, $|\{T_{k}\}| \sim O(2^k)$, the state space of
our model is huge even for relatively short sentences so we had to use
a search strategy that prunes it. Our choice was a synchronous
multi-stack search algorithm which is very similar to a beam search. 

Each stack contains hypotheses --- partial parses --- that have
been constructed by \emph{the same number of predictor and the same number of parser
operations}. The hypotheses in each stack are ranked according to the
$\ln(P(W,T))$ score, highest on top. 
The width of the search is controlled by two parameters:\\
$\bullet$ the maximum stack depth --- the maximum number of hypotheses
  the stack can contain at any given state;\\ 
$\bullet$ log-probability threshold --- the difference between the log-probability score of the top-most
  hypothesis and the bottom-most hypothesis at any given state of the
  stack cannot be larger than a given threshold.

\begin{figure}
  \begin{center} 
    \epsfig{file=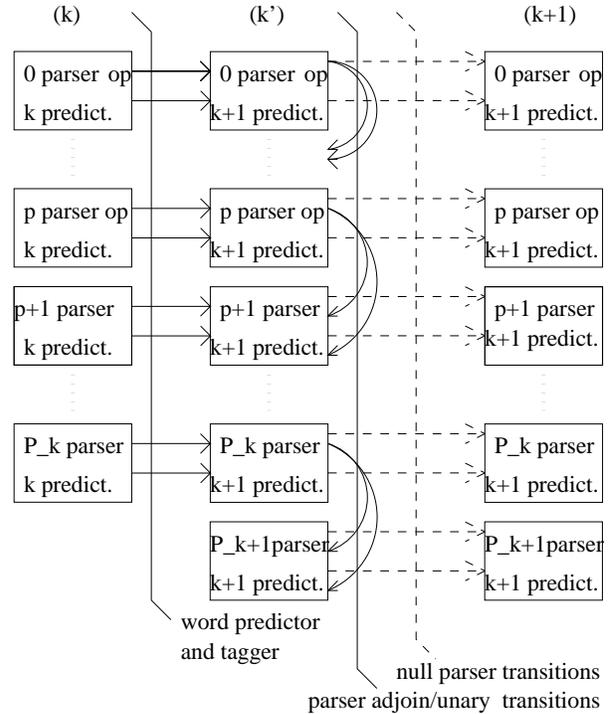,width=\columnwidth}
  \end{center}
  \caption{One search extension cycle} \label{fig:search}
\end{figure}

Figure~\ref{fig:search} shows schematically the operations associated
with the scanning of a new word $w_{k+1}$. 
The above pruning strategy proved to be insufficient so
 we chose to also discard all hypotheses whose score is more than the
log-probability threshold below the score of the topmost
hypothesis. This additional pruning step is performed after all
hypotheses in stage $k'$ have been extended with the \verb+null+
parser transition and thus prepared for scanning a new word.

\subsection{Word Level Perplexity}\label{section:word_level_ppl}

The conditional perplexity calculated by assigning to a whole sentence
the probability:
\begin{eqnarray}
  P(W/T^*) = \prod_{k=0}^n P(w_{k+1}/W_{k}T_{k}^*), \label{eq:ppl0}
\end{eqnarray}
where $T^* = argmax_{T}P(W,T)$, is not valid
because it is not causal: when predicting $w_{k+1}$ we use $T^*$ which was
determined by looking at the entire sentence. To be able to compare 
the perplexity of our model with that resulting from the standard
trigram approach, we need to factor in the entropy of
guessing the correct parse $T_{k}^*$ \emph{before predicting} $w_{k+1}$,
based solely on the word prefix $W_{k}$.

The probability assignment for the word at
position $k+1$ in the input sentence is made using:
\begin{eqnarray}
\lefteqn{P(w_{k+1}/W_{k})=} \nonumber \\
& \sum_{T_{k}\in S_{k}}P(w_{k+1}/W_{k}T_{k})\cdot\rho(W_{k},T_{k}),\label{eq:ppl1}\\ 
\lefteqn{\rho(W_{k},T_{k}) = P(W_{k}T_{k})/\sum_{T_{k} \in S_{k}}P(W_{k}T_{k})} 
\end{eqnarray}
which ensures a proper probability over strings $W^*$, where $S_{k}$ is
the set of all parses present in our stacks at the current stage $k$.

Another possibility for evaluating the word level perplexity of our
model is to approximate the probability of a whole sentence:
\begin{eqnarray}
  P(W) = \sum_{k=1}^N P(W,T^{(k)}) \label{eq:ppl2}
\end{eqnarray}
where $T^{(k)}$ is one of the ``N-best'' --- in the sense defined by
our search --- parses for $W$. This is a deficient probability
assignment, however useful for justifying the model parameter re-estimation.

The two estimates (\ref{eq:ppl1}) and (\ref{eq:ppl2}) are both
consistent in the sense that if the sums are carried over all possible
parses we get the correct value for the word level perplexity of our model. 

\subsection{Parameter Re-estimation}\label{section:model_reest}

The major problem we face when trying to reestimate the model parameters
is the huge state space of the model and the fact that dynamic
programming techniques similar to those used in HMM parameter
re-estimation cannot be used with our model. Our solution is inspired
by an HMM re-estimation technique that works on pruned --- N-best ---
trellises\cite{byrne98}.

Let $(W, T^{(k)}), k = 1 \ldots N$ be the set of hypotheses that
survived our pruning strategy until the end of the parsing process for
sentence $W$. Each of them was produced by a sequence of
model actions, chained together as described in
section~\ref{section:basic_idea}; let us call the sequence of model
actions that produced a given $(W,T)$ the $derivation(W,T)$. 

Let \emph{an elementary event} in the $derivation(W,T)$ be
$(y_l^{(m_l)}, {\underline{x}}_l^{(m_l)})$ where:\\
$\bullet$ $l$ is the index of the current model action;\\
$\bullet$ $m_l$ is the model component --- WORD-PREDICTOR, TAGGER,
PARSER --- that takes action number $l$ in the $derivation(W,T)$;\\
$\bullet$ $y_l^{(m_l)}$ is the action taken at position $l$ in the
derivation:\\ if $m_l$ = WORD-PREDICTOR, then $y_l^{(m_l)}$ is a word;\\
if $m_l$ = TAGGER, then $y_l^{(m_l)}$ is a POStag;\\
if $m_l$ = PARSER, then $y_l^{(m_l)}$ is a parser-action;\\
$\bullet$ ${\underline{x}}_l^{(m_l)}$ is the context in which the above action was
taken: \\
if $m_l$ = WORD-PREDICTOR or PARSER, then 
\mbox{${\underline{x}}_l^{(m_l)} = (h_{0}.tag, h_{0}.word, h_{-1}.tag, h_{-1}.word)$}; \\
if $m_l$ = TAGGER, then \\ \mbox{${\underline{x}}_l^{(m_l)} = ($word-to-tag$, h_{0}.tag, h_{-1}.tag)$}.

The probability associated with each model action is determined as described
in section~\ref{subsection:modeling_tools}, based on counts
$C^{(m)}(y^{(m)}, {\underline{x}}^{(m)})$, one set for each model component.

Assuming that the deleted interpolation coefficients and the count ranges
used for tying them stay fixed, these counts are the only parameters to be re-estimated in
an eventual re-estimation procedure; indeed, once a set of counts
$C^{(m)}(y^{(m)}, {\underline{x}}^{(m)})$ is specified for a given model $m$, we can
easily calculate:\\
$\bullet$ the relative frequency estimates\\
\mbox{$f_n^{(m)}(y^{(m)}/{\underline{x}}_n^{(m)})$} for all context orders \mbox{$n
  = 0 \ldots $maximum-order$(model(m))$};\\
$\bullet$ the count $C^{(m)}({\underline{x}}_n^{(m)})$ used for determining the
$\lambda({\underline{x}}_n^{(m)})$ value to be used with the order-$n$
context ${\underline{x}}_n^{(m)}$.\\
This is all we need for calculating the probability of an
\emph{elementary event} and then the probability of an entire
derivation.

One training iteration of the re-estimation procedure we propose is
described by the following algorithm:
\begin{verbatim}
N-best parse development data; // counts.Ei
// prepare counts.E(i+1) 
for each model component c{
  gather_counts development model_c;
}
\end{verbatim}
In the parsing stage we retain for each ``N-best'' hypothesis $(W, T^{(k)}), k = 1 \ldots N,$
only the quantity\\
$\phi(W,T^{(k)}) = P(W,T^{(k)})/\sum_{k=1}^N P(W,T^{(k)})$ \\
and its $derivation(W,T^{(k)})$. 
We then scan all the derivations in the ``development set'' and,
for each occurrence of the elementary event $(y^{(m)}, {\underline{x}}^{(m)})$ in $derivation(W,T^{(k)})$
we accumulate the value $\phi(W,T^{(k)})$ in the $C^{(m)}(y^{(m)}, {\underline{x}}^{(m)})$ counter
to be used in the next iteration.

The intuition behind this procedure is that $\phi(W,T^{(k)})$ is an
approximation to the $P(T^{(k)}/W)$ probability which places all its
mass on the parses that survived the parsing process; the
above procedure simply accumulates the expected values of the
counts $C^{(m)}(y^{(m)}, {\underline{x}}^{(m)})$ under the $\phi(W,T^{(k)})$
conditional distribution. 
As explained previously, the $C^{(m)}(y^{(m)}, {\underline{x}}^{(m)})$
counts are the parameters defining our model, making our procedure
similar to a rigorous EM approach~\cite{em77}.

A particular --- and very interesting --- case is that of events which
had count zero but get a non-zero count in the next iteration, caused
by the ``N-best'' nature of the re-estimation process. Consider a given
sentence in our ``development'' set. The ``N-best'' derivations for
this sentence are trajectories through the state space of our model. 
They will change from one iteration to the other due to the smoothing
involved in the probability estimation and the change of the
parameters --- event counts --- defining our model, thus allowing new
events to appear and discarding others through purging low
probability events from the stacks. The higher the number of
trajectories per sentence, the more dynamic this change is expected to
be. 

The results we obtained are presented in the experiments section. All
the perplexity evaluations were done using the left-to-right formula
(\ref{eq:ppl1}) (L2R-PPL)
for which the perplexity on the ``development set'' is not guaranteed
to decrease from one iteration to another. However, we believe that
our re-estimation method should not increase the approximation to
perplexity based on (\ref{eq:ppl2}) (SUM-PPL) --- again, on the ``development
set''; we rely on the consistency property outlined at the end of
section~\ref{section:word_level_ppl} to correlate the desired decrease in
L2R-PPL with that in SUM-PPL.  No claim can be made about the change
in either L2R-PPL or SUM-PPL on test data.

\subsection{Initial Parameters}\label{section:initial_parameters}

Each model component --- WORD-PREDICTOR, TAGGER, PARSER ---
is trained initially from a set of parsed sentences, after
each parse tree $(W,T)$ undergoes:\\
$\bullet$ headword percolation and binarization --- see section~\ref{section:headword_percolation};\\
$\bullet$ decomposition into its $derivation(W,T)$.\\
Then, separately for each $m$ model component, we:\\
$\bullet$ gather joint counts $C^{(m)}(y^{(m)}, {\underline{x}}^{(m)})$ from the derivations that
make up the ``development data'' using $\phi(W,T) = 1$;\\
$\bullet$ estimate the deleted interpolation coefficients on joint
counts gathered from ``check data'' using the EM algorithm.\\
These are the initial parameters used with the re-estimation procedure
described in the previous section.

\section{Headword Percolation and Binarization} \label{section:headword_percolation}

In order to get initial statistics for our model components we needed
to binarize the UPenn~Treebank~\cite{Upenn} parse trees and percolate
headwords. The procedure we used was to first percolate headwords
using a context-free (CF) rule-based approach and then binarize the parses
by using a rule-based approach again.

The headword of a phrase is the word that best represents the phrase,
all the other words in the phrase being modifiers of the
headword. Statistically speaking, we were satisfied with the output of
an enhanced version of the  procedure described in~\cite{mike96} --- also known under the name
``Magerman \& Black Headword Percolation Rules''. 

\begin{figure}
  \begin{center} 
    \epsfig{file=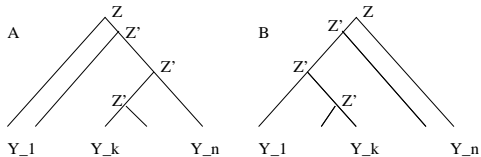,height=2cm,width=6cm}
  \end{center}
  \caption{Binarization schemes} \label{fig:bin_schemes}
\end{figure}

Once the position of the headword within a constituent
--- equivalent with a CF production of the type 
$Z \rightarrow Y_1 \ldots Y_n$ , where $Z, Y_1, \ldots Y_n$ are
non-terminal labels or POStags (only for $Y_i$) ---  is identified to be $k$, we binarize
the constituent as follows: depending on the $Z$ identity, a fixed rule is used to decide
which of the two binarization schemes in Figure~\ref{fig:bin_schemes} to apply.
The intermediate nodes created by the above binarization schemes
receive the non-terminal label $Z'$. 

\section{Experiments} \label{section:experiments}

Due to the low speed of the parser --- 200 wds/min for stack depth 10 and log-probability
threshold 6.91 nats (1/1000) --- we could carry out the re-estimation technique
described in section~\ref{section:model_reest} on only 1 Mwds
of training data. For convenience we chose to work on the UPenn
Treebank corpus. 
The vocabulary sizes were:\\
$\bullet$ word vocabulary: 10k, open --- all words outside the
  vocabulary are mapped to the \verb+<unk>+ token;\\
$\bullet$ POS tag vocabulary: 40, closed;\\
$\bullet$ non-terminal tag vocabulary: 52, closed;\\
$\bullet$ parser operation vocabulary: 107, closed;\\
The training data was split into ``development'' set --- 929,564wds
(sections 00-20) --- and ``check set'' --- 73,760wds (sections 21-22); the
test set size was 82,430wds (sections 23-24). The ``check'' set has been
used for estimating the interpolation weights and tuning the search
parameters; the ``development'' set has been used for
gathering/estimating counts; the test set has been used strictly for
evaluating model performance. 

Table~\ref{table:reest_ppls} shows the results of the re-estimation technique
presented in section~\ref{section:model_reest}. We achieved a reduction
in test-data perplexity bringing an improvement over a deleted
interpolation trigram model whose perplexity was 167.14 on the same
training-test data; the reduction is statistically significant according to a sign test. 
\begin{center}
  \begin{table}[h]
     \begin{center}
       \begin{tabular}{||c|c|c||} \hline
         iteration & DEV set & TEST set      \\ 
         number    & L2R-PPL & L2R-PPL       \\ \hline \hline
         E0        & 24.70   &  167.47       \\ \hline 
         E1        & 22.34   &  160.76       \\ \hline 
         E2        & 21.69   &  158.97       \\ \hline 
         E3        & 21.26   &  158.28       \\ \hline \hline
         3-gram    & 21.20   &  167.14       \\ \hline 
       \end{tabular}
     \end{center}
     \caption{Parameter re-estimation results} \label{table:reest_ppls}
   \end{table}
\end{center}

Simple linear interpolation between our model and the trigram model:
\begin{eqnarray*}
\lefteqn{Q(w_{k+1}/W_k) =} \\
& \lambda \cdot P(w_{k+1}/w_{k-1}, w_k) + (1 - \lambda) \cdot P(w_{k+1}/W_k) 
\end{eqnarray*} 
yielded a further improvement in PPL, as shown in
Table~\ref{table:interpolated_PPL}. The interpolation weight was
estimated on check data to be $\lambda = 0.36$.
\begin{center}
  \begin{table}[h]
     \begin{center}
       \begin{tabular}{||c|c|c||} \hline
         iteration & TEST set & TEST set\\ 
         number    & L2R-PPL  & 3-gram interpolated PPL   \\ \hline \hline
         E0        &  167.47  & 152.25  \\ \hline 
         E3        &  158.28  & 148.90   \\ \hline \hline 
         3-gram    &  167.14  & 167.14       \\ \hline 
       \end{tabular}
     \end{center}
     \caption{Interpolation with trigram results} \label{table:interpolated_PPL}
   \end{table}
\end{center}
An overall relative reduction of 11\% over the trigram model has been
achieved.

\section{Conclusions and Future Directions}

The large difference between the perplexity of our model calculated on
the ``development'' set --- used for model parameter estimation  ---
and ``test'' set --- unseen data --- shows that the initial point we
choose for the parameter values has already captured a lot of
information from the training data. The same problem is encountered in
standard n-gram language modeling; however, our approach has more
flexibility in dealing with it due to the possibility of reestimating
the model parameters.

We believe that the above experiments show the potential
of our approach for improved language models.
Our future plans include:\\
$\bullet$ experiment with other parameterizations than the two most
recent exposed heads in the word predictor model and parser;\\
$\bullet$ estimate a separate word predictor for left-to-right
language modeling. Note that the corresponding model predictor was
obtained via re-estimation aimed at increasing the probability of the
"N-best" parses of the entire sentence;\\
$\bullet$ reduce vocabulary of parser operations; extreme case: no
non-terminal labels/POS tags, word only model; this will increase the
speed of the parser thus rendering it usable on larger amounts of
training data and allowing the use of deeper stacks --- resulting in
more ``N-best'' derivations per sentence during re-estimation;\\
$\bullet$ relax --- flatten --- the initial statistics in the
re-estimation of model parameters; this would allow the model
parameters to converge to a different point that might yield a lower
word-level perplexity;\\
$\bullet$ evaluate model performance on n-best sentences output by an
automatic speech recognizer.

\section{Acknowledgments}
This research has been funded by the NSF \\IRI-19618874 grant (STIMULATE).

The authors would like to thank to Sanjeev Khudanpur for his
insightful suggestions. Also to Harry Printz, Eric Ristad, Andreas
Stolcke, Dekai Wu and all the other members of the dependency modeling
group at the summer96 DoD Workshop for useful comments on the model,
programming support and an extremely creative environment. Also thanks
to Eric Brill, Sanjeev Khudanpur, David Yarowsky, Radu Florian, Lidia
Mangu and Jun Wu for useful input during the meetings of the people
working on our STIMULATE grant.
\bibliographystyle{acl}
\bibliography{acl}

\begin{thebibliography}{}

\bibitem[\protect\citename{Byrne \bgroup et al.\egroup }1998]{byrne98}
W.~Byrne, A.~Gunawardana, and S.~Khudanpur.
\newblock 1998.
\newblock Information geometry and {EM} variants.
\newblock Technical Report CLSP Research Note 17, Department of Electical and
  Computer Engineering, The Johns Hopkins University, Baltimore, MD.

\bibitem[\protect\citename{Chelba \bgroup et al.\egroup }1997]{ws96}
C.~Chelba, D.~Engle, F.~Jelinek, V.~Jimenez, S.~Khudanpur, L.~Mangu, H.~Printz,
  E.~S. Ristad, R.~Rosenfeld, A.~Stolcke, and D.~Wu.
\newblock 1997.
\newblock Structure and performance of a dependency language model.
\newblock In {\em Proceedings of Eurospeech}, volume~5, pages 2775--2778.
  Rhodes, Greece.

\bibitem[\protect\citename{Collins}1996]{mike96}
Michael~John Collins.
\newblock 1996.
\newblock A new statistical parser based on bigram lexical dependencies.
\newblock In {\em Proceedings of the 34th Annual Meeting of the Association for
  Computational Linguistics}, pages 184--191. Santa Cruz, CA.

\bibitem[\protect\citename{Dempster \bgroup et al.\egroup }1977]{em77}
A.~P. Dempster, N.~M. Laird, and D.~B. Rubin.
\newblock 1977.
\newblock Maximum likelihood from incomplete data via the {EM} algorithm.
\newblock In {\em Journal of the Royal Statistical Society}, volume~39 of {\em
  B}, pages 1--38.

\bibitem[\protect\citename{Jelinek and Mercer}1980]{jelinek80}
Frederick Jelinek and Robert Mercer.
\newblock 1980.
\newblock Interpolated estimation of markov source parameters from sparse data.
\newblock In E.~Gelsema and L.~Kanal, editors, {\em Pattern Recognition in
  Practice}, pages 381--397.

\bibitem[\protect\citename{Jelinek \bgroup et al.\egroup }1994]{spatter}
F.~Jelinek, J.~Lafferty, D.~M. Magerman, R.~Mercer, A.~Ratnaparkhi, and
  S.~Roukos.
\newblock 1994.
\newblock Decision tree parsing using a hidden derivational model.
\newblock In ARPA, editor, {\em Proceedings of the Human Language Technology
  Workshop}, pages 272--277.

\bibitem[\protect\citename{Marcus \bgroup et al.\egroup }1995]{Upenn}
M.~Marcus, B.~Santorini, and M.~Marcinkiewicz.
\newblock 1995.
\newblock Building a large annotated corpus of {English}: the {Penn}
  {Treebank}.
\newblock {\em Computational Linguistics}, 19(2):313--330.

\bibitem[\protect\citename{Philips}1996]{colin96}
Colin Philips.
\newblock 1996.
\newblock {\em Order and Structure}.
\newblock {Ph.D.} thesis, MIT.
\newblock Distributed by MITWPL.

\end{thebibliography}
\end{document}